# Witscript 2: A System for Generating Improvised Jokes Without Wordplay

**Joe Toplyn**
Twenty Lane Media, LLC
P. O. Box 51
Rye, NY 10580 USA
joetoplyn@twentylanemedia.com

## Abstract

A previous paper presented Witscript, a system for generating conversational jokes that rely on wordplay. This paper extends that work by presenting Witscript 2, which uses a large language model to generate conversational jokes that rely on common sense instead of wordplay. Like Witscript, Witscript 2 is based on joke-writing algorithms created by an expert comedy writer. Human evaluators judged Witscript 2's responses to input sentences to be jokes 46% of the time, compared to 70% of the time for human-written responses. This is evidence that Witscript 2 represents another step toward giving a chatbot a humanlike sense of humor.

## Introduction

To be truly enjoyable and credible, a conversational agent like a social robot needs to produce contextually integrated jokes about what's happening at the moment (Ritschel, Aslan, Sedlbauer, and André 2019).

Existing computational humor systems can generate conversational jokes that depend on wordplay. But generating jokes that don't rely on wordplay has proven to be more difficult. Indeed, generating all types of humor is often regarded as an AI-complete problem (Winters 2021). Nevertheless, Witscript 2 is a novel system for automatically generating contextually integrated jokes that are based not on wordplay, but on common sense.

## Related Work

Few systems for the computational generation of verbally expressed humor can generate contextually integrated jokes, such as jokes improvised in a conversation. The term verbally expressed humor is used here to mean humor conveyed in language, as opposed to verbal humor, which is sometimes used to mean humor that depends on wordplay (Ritchie 2000).

The method of Zhang, Liu, Lv, and Luo (2020) generates a punch line given a set-up sentence and relevant world knowledge. But its effectiveness is limited because it does not incorporate explicit humor algorithms. The system of Ritschel et al. (2019) transforms a non-ironic utterance into a humorously ironic version using natural language processing and natural language generation techniques. Zhu (2018) uses search engine query statistics to generate a response to a user's utterance that is humorously improbable given the subject of the utterance. The PUNDA Simple system (Dybala, Ptaszynski, Higuchi, Rzepka, and Araki 2008) and the Witscript system (Toplyn 2021) generate joke responses in a conversation, but those jokes rely on wordplay.

In contrast, the Witscript 2 system uses explicit humor algorithms to generate, in real time, conversational jokes that rely on common sense instead of wordplay.

One of those humor algorithms, derived from the Surprise Theory of Laughter (Toplyn 2021), specifies that a monologue-type joke has these three parts:
1. The **topic** is the statement that the joke is based on.
2. The **angle** is a word sequence that smoothly bridges the gap between the topic and the punch line.
3. The **punch line** is the word or phrase that results in a laugh. It's an incongruity at the end of the joke that, surprisingly, turns out to be related to elements of the topic.

Witscript 2, like Witscript, incorporates the Basic Joke-Writing Algorithm (Toplyn 2021), which consists of five steps for writing a three-part joke:
1. **Select a topic.** A good joke topic is one sentence that is likely to capture the attention of the audience.
2. **Select two topic handles.** The topic handles are the two words or phrases in the topic that are the most attention-getting.
3. **Generate associations of the two topic handles.** An association is something that the audience is likely to think of when they think about a particular subject.
4. **Create a punch line.** The punch line links an association of one topic handle to an association of the other topic handle in a surprising way.
5. **Generate an angle between the topic and punch line.** The angle is text that connects the topic to the punch line in a natural-sounding way.

Now I'll describe how the Witscript 2 system executes the five steps of the Basic Joke-Writing Algorithm.

## Description of the Witscript 2 System

The Witscript 2 system is powered by a large language model, OpenAI's GPT-3 (Brown et al. 2020). The most capable GPT-3 model currently available is used, the text-davinci-002 model. GPT-3 was trained on a filtered version of Common Crawl, English-language Wikipedia, and other high-quality datasets. I accessed GPT-3 via the

OpenAI API (https://openai.com/api/) and did not fine-tune the model.

Here's how Witscript 2, using GPT-3 components, executes the Basic Joke-Writing Algorithm to generate a joke response to a conversational topic:

1. **Select a topic.** Witscript 2 receives a sentence from a user and treats it as the topic of a three-part joke. For example, the user tells the system, "The U.S. is planning to buy 22 aging fighter jets from Switzerland."
2. **Select two topic handles.** The GPT-3 API is called with a prompt to select the two most attention-getting nouns, noun phrases, or named entities in the topic. From that example topic, GPT-3 selects the topic handles "fighter jets" and "Switzerland."
3. **Generate associations of the two topic handles.** The GPT-3 API is called with a prompt to generate a list of associations for each topic handle. In our example, for "fighter jets" GPT-3 generates a list including "F-22 Raptor." For "Switzerland" it generates a list including "Swiss chocolate."
4. **Create a punch line.** The GPT-3 API is called with a prompt to select one association from each list and combine them. In our example, GPT-3 selects "F-22 Raptor" and "Swiss chocolate" and combines them to create the punch line "Swiss Chocolate F-22s."
5. **Generate an angle between the topic and punch line.** The GPT-3 API is called with a prompt to generate a joke, based on the topic, that ends with the punch line. In our example, the system generates the joke "I hear they're delicious Swiss Chocolate F-22s." The system responds to the user with the joke.

## System Evaluation

To evaluate Witscript 2, I used the same methodology and 13 input topics that were used to evaluate Witscript (Toplyn 2021). For a new baseline, I used the text-davinci-002 model of GPT-3; for each input topic, the GPT-3 API was called with the prompt "Generate a short response."

I hired workers via Amazon Mechanical Turk (AMT) to evaluate the responses generated by GPT-3 and Witscript 2, and also the responses of Witscript and a human comedy writer that are listed by Toplyn (2021). None of the 13 input topics were used in any prompts given to the GPT-3 API. I specified that the AMT workers had to be located in the United States and have a Human Intelligence Task (HIT) Approval Rate greater than or equal to 95%.

The 52 input and response pairs were put in a random order and presented in that same order to all of the AMT evaluators, who were not told how the pairs had been obtained. Given each input and response pair, the evaluators were asked to rate the response as either 1 (not a joke), 2 (almost a joke), 3 (a joke), or 4 (a very good joke).

Each input and response pair was evaluated by 15 AMT workers (780 ratings total). Table 1 shows the first response that was output by the three systems for each input, along with the human's response. All the responses were transcribed exactly as they were output. The mean rating given by the evaluators for each response is also shown.

| Source | Input and responses | Rating |
|---|---|---|
| | **1. Input: A man pulled out a gun at a Popeyes restaurant when he heard they had no more chicken sandwiches.** | |
| GPT-3 | This is ridiculous! People need to learn to control themselves. | 1.80 |
| Witscript | Popeyes gun? Um, not a Bubba Gun. | 2.53 |
| Witscript 2 | The man was arrested and charged with "attempted Cajun Fried Chicken." | 2.47 |
| Human | He also objected to any new laws on bun control. | 2.40 |
| | **2. Input: America's national forests are being severely damaged by the illegal growing of marijuana.** | |
| GPT-3 | This is a huge problem that needs to be addressed. | 1.53 |
| Witscript | America marijuana? Yah, so now he wants to be not in Canada cannabis. | 2.20 |
| Witscript 2 | The worst part is the people doing it are just "smoking weed in the trees." | 1.73 |
| Human | Which explains the new Forest Service mascot, Tokey the Bear. | 2.33 |
| | **3. Input: For Oktoberfest, Adidas is releasing sneakers that repel beer and vomit.** | |
| GPT-3 | That's pretty slick! | 2.27 |
| Witscript | Adidas vomit? Ah, I like the Puma poop. | 2.80 |
| Witscript 2 | Budweiser Puke. | 2.07 |
| Human | And they're changing their slogan to "Just Spew It." | 2.60 |
| | **4. Input: Two men were arrested in the theft of a hundred thousand diapers.** | |
| GPT-3 | This is a serious crime. If convicted, the two men could face significant jail time. | 1.60 |
| Witscript | Theft diapers? Like, they stole stroller. | 2.00 |
| Witscript 2 | The headline should have read: "Pampers & Robbers." | 2.80 |
| Human | They were charged with potty larceny. | 2.93 |

Table 1: Responses generated for each input, with the mean rating of each response on a scale from 1 (not a joke) to 4 (a very good joke).

|  | **5. Input: It's the twenty-fifth anniversary of the Blue Man Group.** |  |
|---|---|---|
| GPT-3 | That's so cool! | 1.60 |
| Witscript | Blue anniversary? Heh, welcome to the Bluebilee. | 2.73 |
| Witscript 2 | And they're celebrating their Blue Wedding Anniversary. | 2.13 |
| Human | To celebrate, they're starting their own health plan: Blue Man Blue Cross Blue Shield. | 2.87 |
|  | **6. Input: Thirteen TGI Fridays were selling cheap liquor in premium brand bottles.** |  |
| GPT-3 | That's a lot of TGI Fridays. | 1.93 |
| Witscript | Thirteen Fridays? Okay, but I prefer a nine night myself. | 2.67 |
| Witscript 2 | The good news is that they're now serving "Drunken Wings." | 2.47 |
| Human | The company claims it's a new promotion: TGI Fraudays. | 2.73 |
|  | **7. Input: At a zoo, the first rhinoceros ever conceived by artificial insemination was born.** |  |
| GPT-3 | How wonderful! | 1.53 |
| Witscript | Rhinoceros insemination? Mmm-hmm, welcome to the zoosemination. | 1.93 |
| Witscript 2 | The baby rhino's parents are so excited, they can't wait to show him off to their friends. They're going to take him to the in vitro rhinoceros horn. | 2.27 |
| Human | Usually rhinos are conceived using that dating site for large mammals, Ashley Mastodon. | 2.53 |
|  | **8. Input: Dow Chemical is turning raw sewage into drinking water.** |  |
| GPT-3 | Dow Chemical is turning raw sewage into drinking water. This is a great way to recycle water and help save the environment. | 1.87 |
| Witscript | Dow Chemical drinking? Yup, so not Dowing. | 2.33 |
| Witscript 2 | I'm not sure if that's a good thing or a bad thing, but they're calling it the 'Dow Jones Industrial Fountain.' | 1.80 |
| Human | To make it more appealing, they're calling it Pooland Spring. | 3.13 |
|  | **9. Input: A new study says drinking sugar-sweetened beverages increases the risk of death.** |  |
| GPT-3 | This is yet another reason to avoid sugary drinks! | 1.73 |
| Witscript | Beverages risk? Huh, I have to focus on drinks risks. | 2.20 |
| Witscript 2 | The study was sponsored by "Coca-Cola funeral." | 2.07 |
| Human | That's why they're renaming that soft drink Coca-Coma. | 2.53 |
|  | **10. Input: In one of its restrooms, the Guggenheim Museum is installing a solid gold toilet.** |  |
| GPT-3 | This is an interesting bit of news. | 1.80 |
| Witscript | Gold toilet? Yep, to make a toilet pure. | 2.33 |
| Witscript 2 | "The Golden Throne." Yeah, it's a little gaudy, but it's perfect for a museum that's already full of crap. | 2.93 |
| Human | It's perfect for people who have just eaten 24 carrots. | 3.20 |
|  | **11. Input: A brewery apologized for a beer that shares the same name as a Hindu deity.** |  |
| GPT-3 | The brewery has issued an apology and is taking steps to rename the beer. | 1.73 |
| Witscript | Shares Hindu? Heh, no Sikh stock. | 2.07 |
| Witscript 2 | I guess they should have seen that one coming. Now they're getting hammered for "Beer Brahma." | 2.33 |
| Human | No word yet on their plans for their other beer, Buddha-weiser. | 2.73 |
|  | **12. Input: A truck was stolen that contained half a million dollars' worth of handbags.** |  |
| GPT-3 | This is a serious crime with significant financial implications. The truck and its contents should be recovered as soon as possible. | 1.73 |
| Witscript | Truck handbags? Yep, of vanbags. | 2.13 |
| Witscript 2 | The truck was a Ford Vuitton. | 2.73 |
| Human | The thief was arrested and the handbags were charged as accessories. | 3.07 |
|  | **13. Input: A man was caught during a home burglary wearing only lingerie.** |  |
| GPT-3 | This is an outrageous story! | 1.67 |
| Witscript | Burglary lingerie? Yes, and the burglaries bikinis. | 2.47 |
| Witscript 2 | Victoria's Secret police report. | 2.60 |
| Human | He was charged with panty larceny. | 2.93 |

Table 1 (continued).

Table 2 compares the systems based on their ratings. The second column shows that Witscript 2's responses were rated, on average, about halfway between those of the GPT-3 baseline and the human, a professional comedy writer. Witscript 2's responses were also rated, on average, the same as Witscript's responses, a result that may not seem particularly impressive. But that result is encouraging because it shows that Witscript 2 can create jokes that are as successful as, but more sophisticated than, mere word-play jokes, which are usually regarded as the low-hanging fruit of computational humor.

The last column of Table 2 shows the percentage of responses that the evaluators rated as "a joke" or "a very good joke." Witscript 2's responses were judged to be jokes 46% of the time, compared to only 25% of the time for the GPT-3 baseline responses. This result, too, is encouraging because it is additional evidence that a system to generate contextually integrated jokes is feasible.

| System | Mean rating | % jokes (ratings of 3 or 4) |
|---|---|---|
| GPT-3 | 1.75 | 25.1% |
| Witscript | 2.34 | 47.2% |
| Witscript 2 | 2.34 | 46.2% |
| Human | 2.77 | 70.3% |

Table 2: Comparison of the systems based on their ratings.

## Discussion

### Computational Creativity

I believe that the Witscript 2 system demonstrates computational creativity instead of mere generation because its output exhibits three characteristics: novelty, value, and intentionality (Ventura 2016).

The system's output has **novelty** because each contextually relevant joke that the system improvises in response to a new input has almost certainly never been created before by it or by any other agent.

The system's output has **value** in that human evaluators judge the system's responses to be jokes 46% of the time, and conversational jokes like those output by the system have worth and usefulness (Dybala et al. 2008).

And the system produces that novel, valuable output with **intentionality** in several ways: It restricts its generation process by using domain knowledge about how a professionally-written joke is structured. It generates jokes in an autonomous fashion by using a language model prompted with an inspiring set consisting of quality examples, namely professionally-written jokes. Finally, it apparently employs a fitness function to intentionally filter out joke responses that don't meet some threshold of value.

For example, given the topic "Today the Arby's fast food chain announced the release of a vodka that tastes like their French fries," Witscript 2 responded, "The good news is, now you can get drunk and fat at the same time." In doing so, it deliberately rejected the punch line that it had generated using the Basic Joke-Writing Algorithm: "Smirnoff and McDonald's." Instead, it improvised a different punch line and created a joke that it somehow decided was more worthy of being output.

### Commonsense Knowledge

In addition to computational creativity, Witscript 2 demonstrates commonsense knowledge. This commonsense knowledge consists of knowledge of everyday commonsense relations.

For example, in generating the joke about the fighter jets from Switzerland, Witscript 2 exhibits taxonomic reasoning (Davis and Marcus 2015) when it infers that "F-22 Raptor" is an instance of "fighter jets." In terms of the commonsense relation types in the commonsense knowledge graph ATOMIC 2020 (Hwang et al. 2021), Witscript 2 humorously infers that a physical object from Switzerland would be made of ("MadeUpOf ") a material commonly found in ("AtLocation") Switzerland, i.e., Swiss chocolate. Witscript 2 also infers that a physical object made of Swiss chocolate would be ("HasProperty") delicious.

### Contributions

This paper makes the following contributions:
1. It introduces a novel system for automatically improvising contextually integrated jokes that don't depend on wordplay.
2. It shows how computational humor can be implemented with a hybrid of a large language model and symbolic AI, where the symbolic AI incorporates expert knowledge of comedy domain rules and algorithms.
3. It demonstrates that generating humor that relies on some commonsense knowledge may not be an AI-complete problem.

### Future Work

I anticipate that future work will improve the performance of Witscript 2 until its jokes based on common sense are rated better than the wordplay jokes of Witscript. To that end, work will be directed toward getting Witscript 2 to execute the Basic Joke-Writing Algorithm more effectively.

To accomplish that, the following will be explored: using different prompts and configuration settings for base GPT-3 models; fine-tuning base GPT-3 models to create multiple customized versions, each version optimized to carry out one joke-writing step; and substituting different large language models for GPT-3.

## Conclusion

The Witscript 2 joke generation system could be integrated into a chatbot as a humor module; the proprietary software is available for license. Such a humor-enabled chatbot might potentially animate an artificial, but likeable, companion for lonely humans.

## Acknowledgments

The author would like to thank three anonymous reviewers for their helpful comments.

## Author Contributions

J. T. ideated and wrote the paper alone.